\documentclass[letterpaper, 10 pt, journal, twoside]{IEEEtran}

\usepackage{graphics}
\usepackage{amsmath,amssymb}
\usepackage{import}
\usepackage[detect-all=true,separate-uncertainty=true, multi-part-units=single,range-phrase={-},product-units = single]{siunitx}

\usepackage{menukeys}
\usetikzlibrary{arrows,positioning,shapes.geometric}
\usetikzlibrary{intersections}          
\usepackage{amsfonts}
\usepackage[caption=false,font=small]{subfig}
\usepackage{booktabs}
\usepackage{changepage}
\DeclareSIUnit\year{y}
\usepackage{hyperref}
\usepackage{wasysym}
\usepackage{ifthen}
\usepackage{ulem}

\usepackage[switch]{lineno}

\begin{document}
\bstctlcite{IEEEexample:BSTcontrol}

\title{A Digital Twin for Robotic Post Mortem Tissue Sampling using Virtual Reality}

\author{Maximilian~Neidhardt$^{1,3,4}$*,
Ludwig~Bosse$^{1}$*,
Vidas~Raudonis$^{4}$,
Kristina~Allgoewer$^{2}$,
Axel~Heinemann$^{2}$,
Benjamin~Ondruschka$^{2}$ 
and Alexander~Schlaefer$^{1,3,4}$
\thanks{*Both authors contributed equally.}
\thanks{$^{1}$Institute of Medical Technology and Intelligent Systems, Hamburg University of Technology, 21073 Hamburg, Germany}%
\thanks{$^{2}$Institute	of	Legal	Medicine,	University	Medical	Center	Hamburg-Eppendorf, 22529 Hamburg, Germany}%
\thanks{$^{3}$Interdisciplinary Competence Center for Interface Research, 22529 Hamburg, Germany}
\thanks{$^{4}$SustAInLivWork Center of Excellence, Kaunas, Lithuania}
\thanks{This research was partially funded by the European Union under Horizon Europe programme (G/N 101059903), NaviTip project (G/N 16KN113039), the NATON project (funded by the German Federal Ministry of Education and Research BMBF within the Network University Medicine; G/N 01KX2121) and the German Research Foundation (G/N ON184/1-1).}%
\thanks{Ethical approval: The Ethics Committee of the Hamburg Chamber of Physicians approved the study (2020-10353-BO-ff).}%
}

\markboth{IEEE Robotics and Automation Letters. Preprint Version. Accepted August 2025}
{Neidhardt \MakeLowercase{\textit{et al.}}: VR-guided Robotic post mortem Tissue Biopsy} 

\maketitle

\begin{abstract}
Studying tissue samples obtained during autopsies is the gold standard when diagnosing the cause of death and for understanding disease pathophysiology. Recently, the interest in post mortem minimally invasive biopsies has grown which is a less destructive approach in comparison to an open autopsy and reduces the risk of infection. While manual biopsies under ultrasound guidance are more widely performed, robotic post mortem biopsies have been recently proposed. This approach can further reduce the risk of infection for physicians. However, planning of the procedure and control of the robot need to be efficient and usable. We explore a virtual reality setup with a digital twin to realize fully remote planning and control of robotic post mortem biopsies. The setup is evaluated with forensic pathologists in a usability study for three interaction methods. Furthermore, we evaluate clinical feasibility and evaluate the system with three human cadavers. Overall, 132 needle insertions were performed with an off-axis needle placement error of \SI{5.30(3.25)}{\milli\meter}. Tissue samples were successfully biopsied and histopathologically verified. Users reported a very intuitive needle placement approach, indicating that the system is a promising, precise, and low-risk alternative to conventional approaches.
\end{abstract}

\begin{IEEEkeywords}
Virtual Reality, Robot Control, User Interface, Soft Tissue, Path Planning
\end{IEEEkeywords}

\IEEEpeerreviewmaketitle

\section{Introduction}
\IEEEPARstart{T}{issue} samples are important to investigate the underlying pathophysiology and, if sampled post mortem (PM), to investigate the cause of death. Commonly, multiple tissue samples are extracted from a cadaver during an open autopsy. However, its invasive nature entails limitations, e.g., it may be rejected by next of kin and it carries a risk of infection for clinical and forensic pathologists \cite{brandner2022contamination}. For example, during the COVID-19 pandemic authorities were initially hesitant to order autopsies to avoid further spread of the novel virus. Note that consequences of infection are even more serious for other diseases, such as viral hemorrhagic fevers. Hence, minimally invasive tissue sampling has been proposed as an alternative to conventional autopsies in studying the pathomechanisms of infectious diseases \cite{brandner2022contamination}, \cite{lahmer2024postmortem}.

\begin{figure}[t]
    \centering
     \begin{tikzpicture}
             \node [align=left, anchor=south west] at (0,0)  { \includegraphics[width = 0.98\linewidth,trim=0cm 0cm 0mm 0cm,clip]{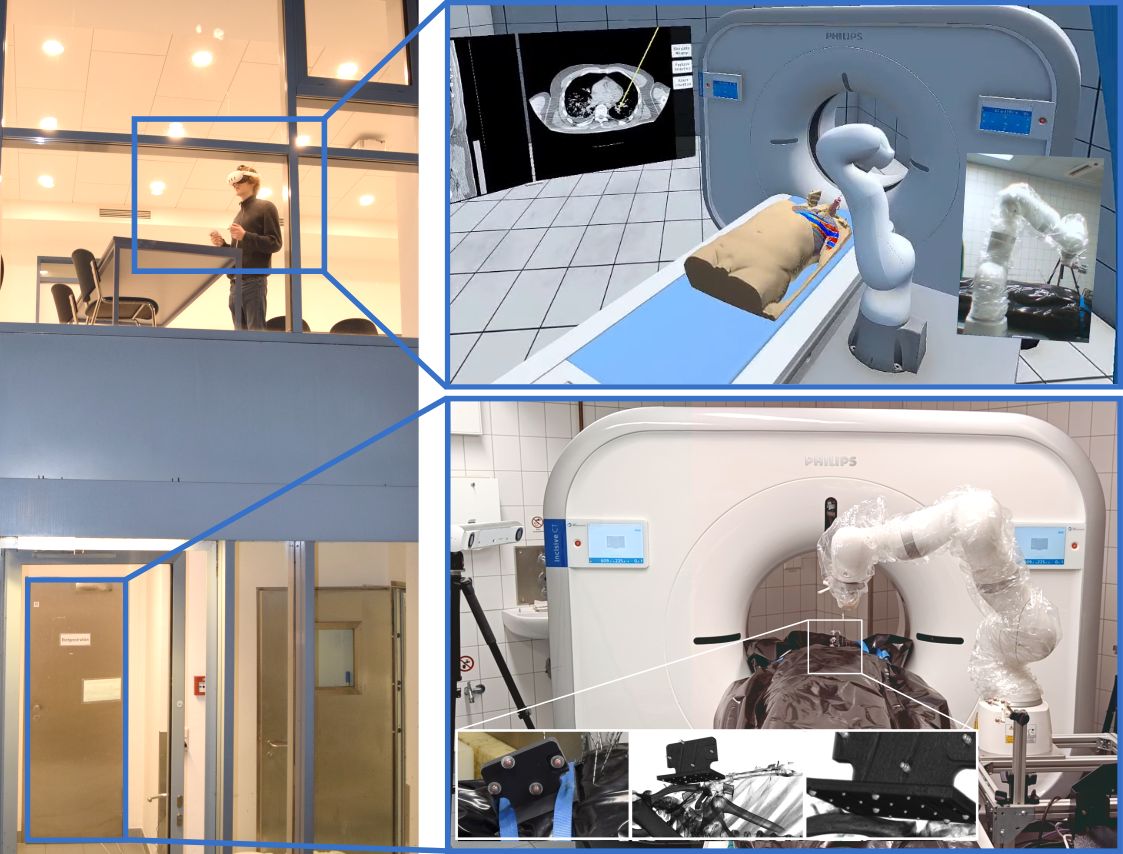}};
             \draw [fill=black] (4,2.6) circle (0.2mm); 
             \draw [] (4,2.6) -- (4-0.1,2.6+0.25)   node [above=-1mm] {\scriptsize E};
             \draw [fill=black] (5.8,2.8) circle (0.2mm); 
             \draw [] (5.8,2.8) -- (5.8-0.25,2.8+0.25)   node [above left=-1mm] {\scriptsize D};
             \draw [fill=black] (6.25,1.5) circle (0.2mm); 
             \draw [] (6.25,1.5) -- (6.25-0.6,1.5+0.6)   node [above left=-1mm] {\scriptsize C};
             \draw [fill=black] (7.5,2.8) circle (0.2mm); 
             \draw [] (7.5,2.8) -- (7.5+0.25,2.8+0.2)   node [above right=-1mm] {\scriptsize A};
             
             \draw [fill=black] (8.7,0.86) circle (0.2mm); 
             \draw [] (8.7,0.86) -- (8.7-0.2,0.86+0.28)   node [above=-1mm] {\scriptsize B};
             
             \draw [fill=black] (5.1,1.0) circle (0.2mm); 
             \draw [] (5.1,1.0) -- (5.1+0.2,1.0+0.15)   node [above right=-1mm] {\scriptsize F};

            \fill[fill=white, opacity=0.7] (0.122,6.728) rectangle (3.36,6.725 - 0.4);
            \fill[fill=white, opacity=0.7] (8.76-3.235,6.686) rectangle (8.76,6.686 - 0.4);
            \fill[fill=white, opacity=0.7] (8.76-3.235,3.625) rectangle (8.76,3.625 - 0.4);
            
            \node[align=center] at (0.125 + 3.235/2,6.55) {\scriptsize Institute of Legal Medicine};
             
            \node[align=center] at (8.76 - 3.235/2,6.48) {\scriptsize Virtual Reality Application};
            
            \node[align=center] at (8.76 - 3.235/2,3.43) {\scriptsize Intervention Room};
               
        \end{tikzpicture}
    \caption{\textbf{Virtual reality guided tissue biopsy with a robot:} Top: The user defines biopsy targets in CT images and plans 3-dimensional insertion paths inside a custom-designed VR application. The system allows fully remote planning and control of robotic biopsies. Bottom: An LBR Med robot (A) mounted to a mobile platform (B) positions needles inside a cadaver (C). Before insertions, a CT scan (D) is acquired. The robot is calibrated relative to the CT system with a tracking camera (E) and a phantom (F). The clinician is not present in the intervention room during planning and needle insertion.}
    \label{fig:setup}
\end{figure}
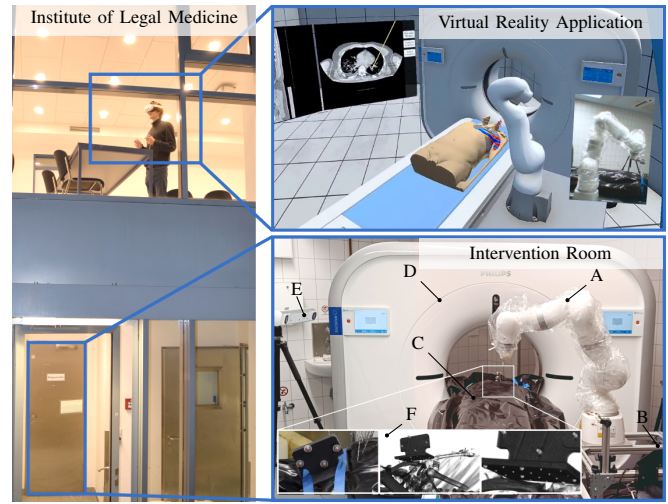

\begin{figure*}[bt]
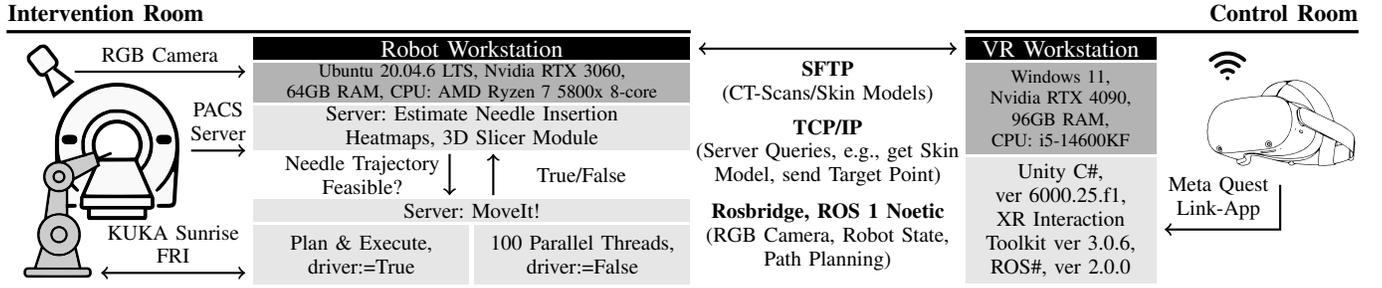

    \centering
    \include{communicationFigure.tex}
    \vspace{-12mm}
     \caption{\textbf{System communication:} Overview of the communication links between the control and the intervention room. 
     }
    \label{fig:systemCommunication}
\end{figure*}
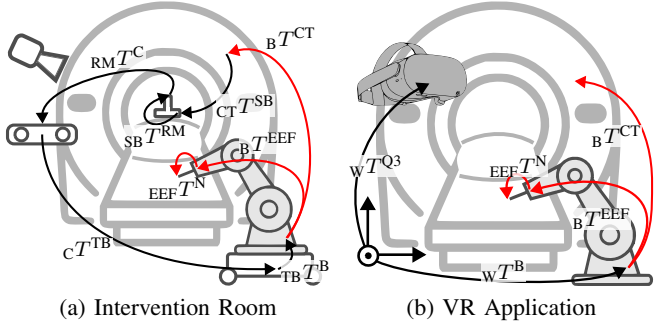
\begin{figure}[t]
    \centering
        \subfloat[Intervention Room]{
        \begin{tikzpicture}[every node/.style={inner sep=0,outer sep=0}]
            \node [align=left, anchor=south west] at (0,0)  {\includegraphics[height = 38mm,trim=0cm 0cm 0cm 0cm,clip]{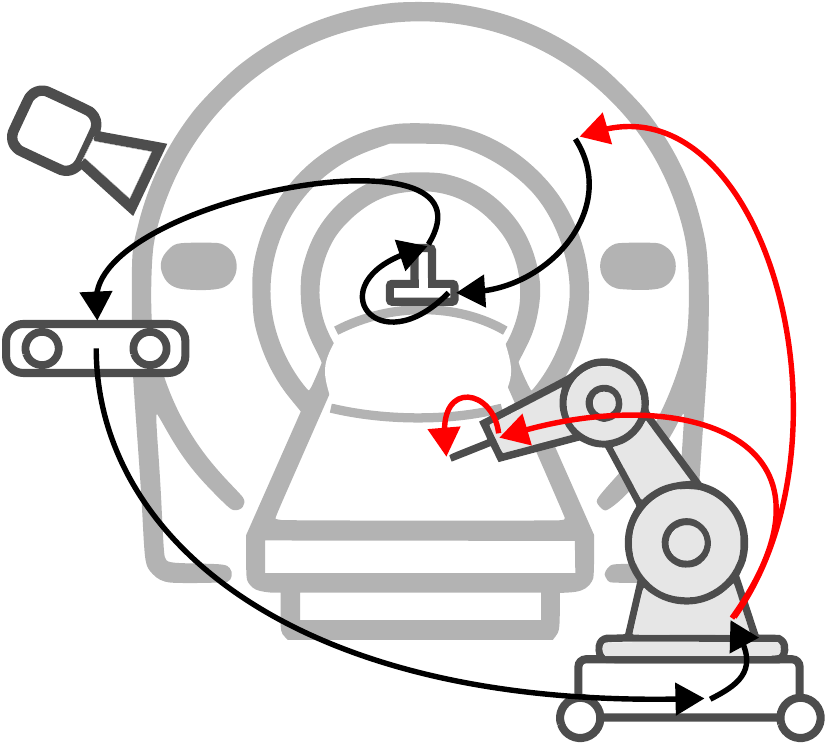}};
          
            \node[align=left,fill=black!0,opacity=.8,text opacity=1] at (3.8,3.3) {\footnotesize{$_{\text{B}}T^{\text{CT}}$}};
            \node[align=left,fill=black!0,opacity=.8,text opacity=1] at (3.2,2.4) {\footnotesize{$_{\text{CT}}T^{\text{SB}}$}};
            \node[align=left,fill=black!0,opacity=.8,text opacity=1] at (2.0,2.0) {\footnotesize{$_{\text{SB}}T^{\text{RM}}$}};
            \node[align=left,fill=black!0,opacity=.8,text opacity=1] at (1.5,3.05) {\footnotesize{$_{\text{RM}}T^{\text{C}}$}};
            \node[align=left,fill=black!0,opacity=.8,text opacity=1] at (1.1,0.5) {\footnotesize{$_{\text{C}}T^{\text{TB}}$}};
            \node[align=left,fill=black!0,opacity=.8,text opacity=1] at (4.0,0.2) {\footnotesize{$_{\text{TB}}T^{\text{B}}$}};

            \node[align=left,fill=black!0,opacity=.8,text opacity=1] at (3.5,1.9) {\footnotesize{$_{\text{B}}T^{\text{EEF}}$}};
            \node[align=left,fill=black!0,opacity=.8,text opacity=1] at (2.3,1.3) {\footnotesize{$_{\text{EEF}}T^{\text{N}}$}};

        \label{fig:transformsB}
        \end{tikzpicture}    
        }
        \subfloat[VR Application]{
        \begin{tikzpicture}[every node/.style={inner sep=0,outer sep=0}]
            \node [align=left, anchor=south west] at (0,0)  {\includegraphics[height = 38mm,trim=0cm 0cm 0cm 0cm,clip]{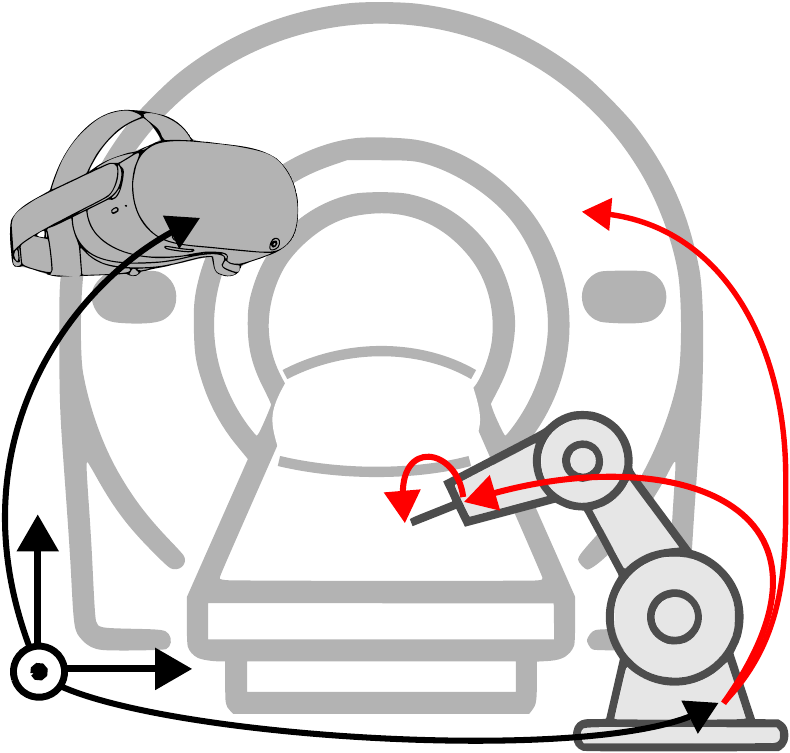}};
          
            \node[align=left,fill=black!0,opacity=.8,text opacity=1] at (2.0,0.2) {\footnotesize{$_{\text{W}}T^{\text{B}}$}};
            \node[align=left,fill=black!0,opacity=.8,text opacity=1] at (3.5,2.0) {\footnotesize{$_{\text{B}}T^{\text{CT}}$}};
            \node[align=left,fill=black!0,opacity=.8,text opacity=1] at (3.3,0.9) {\footnotesize{$_{\text{B}}T^{\text{EEF}}$}};
            \node[align=left,fill=black!0,opacity=.8,text opacity=1] at (2.2,1.6) {\footnotesize{$_{\text{EEF}}T^{\text{N}}$}};
            \node[align=left,fill=black!0,opacity=.8,text opacity=1] at (0.3,1.6) {\footnotesize{$_{\text{W}}T^{\text{Q3}}$}};

        \label{fig:transformsA}
        \end{tikzpicture}
        }  
         \caption{\textbf{Relevant transformations:} System transforms applied in the intervention room (a) and in the VR application (b). Transforms indicated in red are identical for both.}
 \label{fig:transforms}
\end{figure}
Typically, during minimally invasive tissue sampling, the physician manually inserts a needle close to the tissue target using image guidance, e.g., by ultrasound \cite{terence2023minimally}. While commonly applied in clinical practice, manual PM biopsies pose additional challenges. First, direct contact between the performing pathologist and the cadaver should be minimized to limit potential disease transmission. Second, needle placement by hand can be tedious, especially when systematically sampling a large number of tissue samples, e.g., for biobanking during a pandemic. Inserting biopsy needles with a robot can compensate for these drawbacks.
Several robotic systems for pre-aligning a needle have been proposed \cite{levy2021clinical,bodard2023percutaneous}. Systems that visually assist the surgeon during needle placement by hand are used clinically, e.g., with a laser mounted on a robot projecting the needle trajectory \cite{moser2013novel}. To further assist the physicians, collaborative robotic needle insertion has been studied with haptic force feedback from needle tip forces \cite{mieling2023collaborative}. In contrast to clinical biopsies, PM tissue sampling offers greater potential for automation, as vital structures can be penetrated and do not continuously translate in the absence of respiratory motion. To this end, a fixed ceiling-mounted robotic system for inserting biopsy needles has been presented \cite{Ebert.2010}. More recently, a flexible approach using a lightweight robotic arm has been demonstrated \cite{neidhardt2022robotic}. This system can be fully deployed and removed from the interventional suite. 

The examples illustrate that automatic needle placement in cadavers is feasible. However, another aspect is the integration of robotic systems in the clinical workflow. Ideally, interaction and path planning should be simple and tractable for physicians, without requiring extensive training. Particularly, virtual reality (VR) and augmented reality (AR) allow planning of robotic needle insertions. Several AR-based systems that project clinical image data directly onto the patient during needle placement exist \cite{evans2025augmented}. Also, VR systems for clinical diagnostics \cite{murphy2024diagnostic} and to enable 3D planning of surgical interventions, e.g., for pedicle screw placement \cite{pinter2020slicervr}, have been studied. These systems typically reduce the cognitive load when interacting with 3D structures and offer a more intuitive representation of volumetric image data. Also, remote operation of a robot is feasible, but has not been studied for PM tissue biopsies.

In this work, we present a novel mobile VR-guided robotic PM biopsy system. We designed a VR intervention room as a digital twin, including a robot for actual needle insertion. Medical experts interact with the CT data on a life-size model and directly define the target and the needle path without physical presence in the intervention room. This intuitive planning approach includes robot motion constraints and mitigates potential disease transmission to medical personnel. First, we summarize the proposed system. Second, we introduce three VR planning scenarios with different levels of immersion. Third, we evaluate user interaction with a cohort of medical experts. Finally, we extract actual tissue samples from three cadavers and estimate the sampling accuracy.

\section{Methods}
\subsection{System Components and Calibration}
The system is set up in two separate rooms. As shown in Fig.~\ref{fig:setup}, the user is located in the control room while the cadaver, robot, and CT scanner are located in the intervention room. The user is equipped with a commercially available VR headset (Quest3, Meta, California, USA) with a display resolution of \qtyproduct{2064 x 2208}{pixels} per eye and a frame rate of \SI{72}{\Hz}. The setup in the intervention room is depicted in Fig.~\ref{fig:setup}, bottom. Volumetric images of the cadaver are acquired with a CT system (Incisive CT 128, Philips, Hamburg, Germany) with a slice resolution of \SI{0.45}{\milli\meter} and an in-slice spatial resolution of \SI{0.98}{\milli\meter}. A lightweight robot (LBR Med 14, KUKA, Augsburg, Germany) is mounted on a mobile platform that can be freely positioned next to the CT table. The robot is equipped with a custom end-effector (EEF) to mount biopsy needles. An optical tracking camera with reference \mbox{system $C$} (fusionTrack 500, Atracsys LLC, Puidoux, Switzerland) records the positions of retro-reflective markers attached to the EEF. Also, a reference \mbox{system $TB$} is defined on the mobile platform. A custom phantom with retro-reflective markers $R\!M$ and embedded steel balls $S\!B$ is positioned on the thorax of the cadaver and fixed with two tension straps, similar to \cite{neidhardt2022robotic}. To allow robust estimates $S\!B$ from CT image data, a CNC-machined grid (\qtyproduct{5 x 4}{}, \qtyproduct{25 x 30}{\milli\meter} spacing) is integrated in the phantom with steel balls of radii \SI{2}{\milli\meter} and \SI{5}{\milli\meter} to facilitate pose estimation. The rigid transformation between the retro-reflective marker and the steel balls $_{\text{RM}}T^{\text{SB}}$ is estimated with a calibrated stylus that is tracked by the camera $C$. The rigid transformation between the $E\!E\!F$ and the needle tip is calibrated with a four-point pivot calibration method (Sunrise.OS Med 2.6, KUKA, Augsburg, Germany).

\begin{figure}[t]
    \centering
          \begin{tikzpicture}
            \node [anchor=south west] at (-1.0mm,-1mm)  {\includegraphics[width = 52mm,trim=0mm 0cm 0mm 0cm,clip]{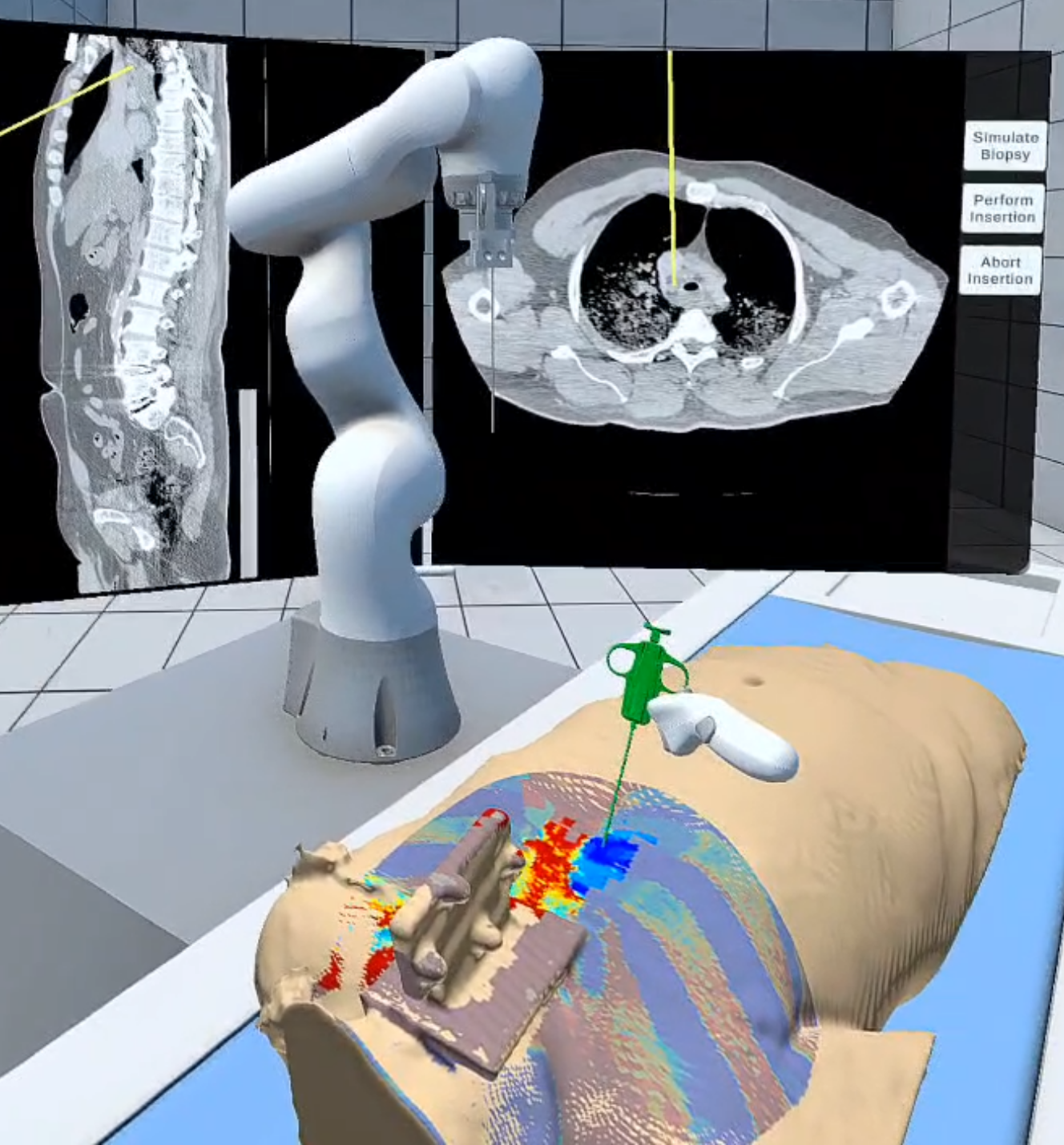}};
            \node [anchor=south west] at (52mm,-1mm)  {\includegraphics[width = 32mm,trim=0mm 0cm 0mm 0cm,clip]{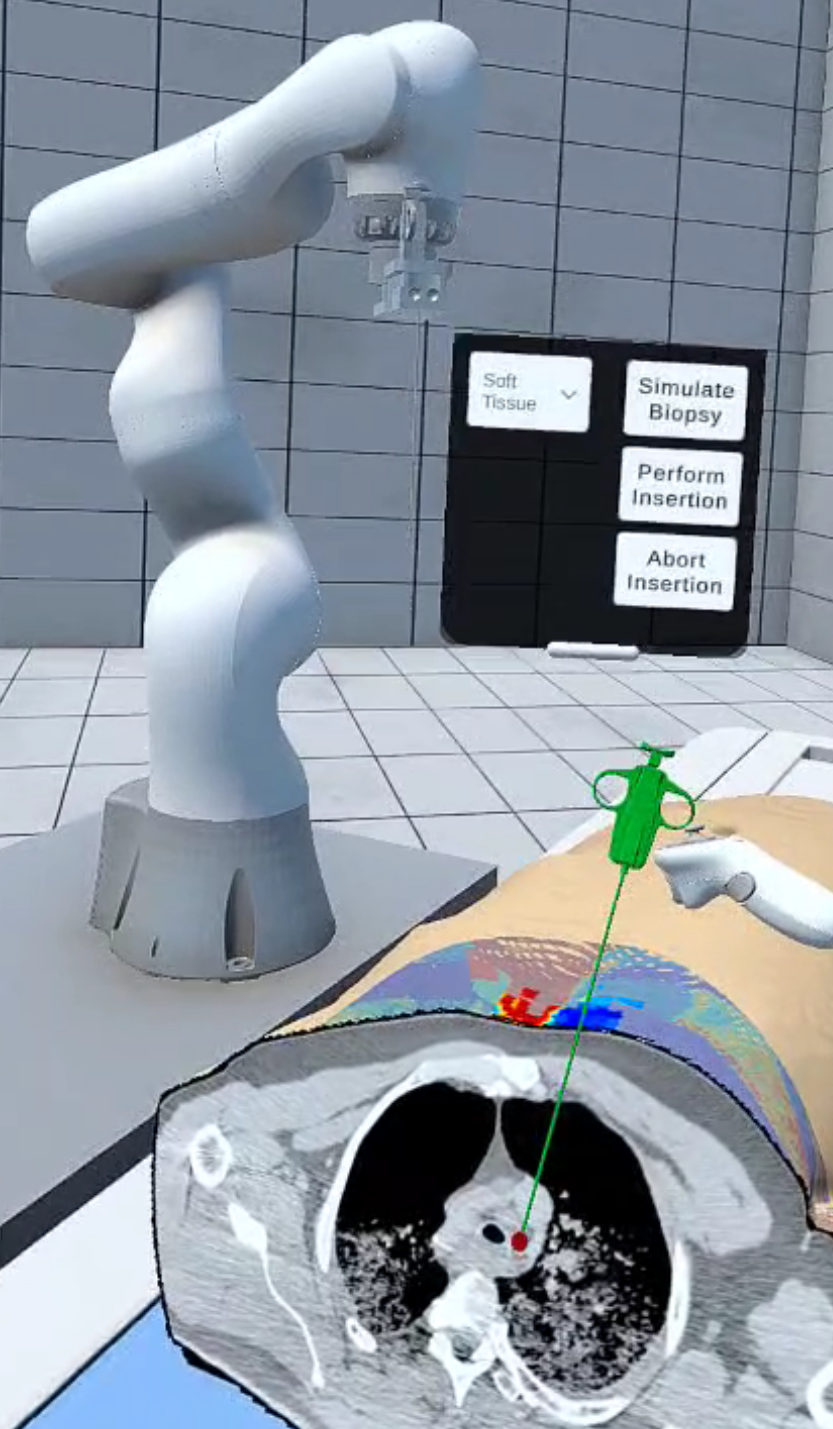}};  
            \fill[fill=white] (0mm,52mm) rectangle (85mm,55mm);
            \node [anchor=south west] at (-1mm,52mm)  {\includegraphics[width = 85mm,trim=0mm 0cm 0mm 0cm,clip]{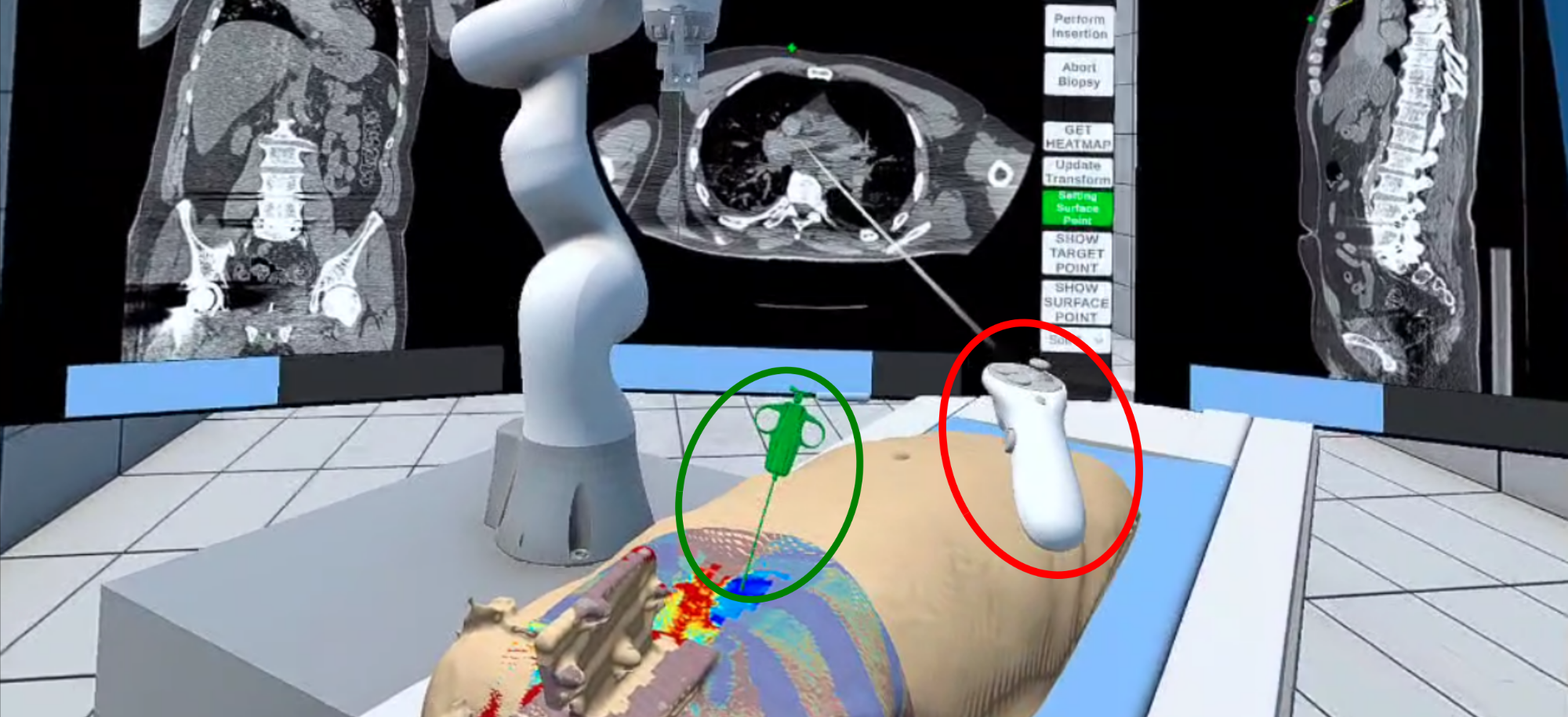}};

             \node[align=center,anchor=north east, fill=white, opacity=0.8] at (86mm,52.1mm) {\scriptsize \textbf{Scenario~3}\\[-1.5mm] \scriptsize 3D Planning};
             \node[align=center,anchor=north east, fill=white, opacity=0.8] at (52.5mm,52.1mm) {\scriptsize \textbf{Scenario~2} \\[-1.5mm] \scriptsize Hybrid};
             \node[align=center,anchor=north east, fill=white, opacity=0.8] at (86mm,92.5mm) {\scriptsize \textbf{Scenario~1} \\[-1.5mm] \scriptsize 2D Planning};
             
            \node [align=left,left=-1mm, fill=black!0,opacity=.8,text opacity=1] at (1.9,0.55){\scriptsize Robot Insertion: \\[-1.5mm] \scriptsize Feasible Area\\[-1.5mm] \scriptsize Unfeasible Area};
            
            \draw [fill=black] (3,1.4) circle (0.2mm); 
            \draw [] (1.1+0.65,0.55) -- (3,1.4);
                    
            \draw [fill=black] (3.2,0.6) circle (0.2mm); 
            \draw [] (1.3+0.65,0.25) -- (3.2,0.6);
                 
            \end{tikzpicture}
          \caption{\textbf{VR needle path planning scenarios:} We design 3 scenarios to plan needle insertion points. Each scenario contains a CT scanner, a screen, and robot. Projected on the skin model is the maximum Hounsfield unit between insertion and target point. Saturated colors indicate insertions that are feasible for the robot and faded colors vice versa. The green biopsy needle (green ellipse) indicates the current chosen needle trajectory. In scenario~1 the user defines the needle insertion point on the screen with a hand-held controller (red ellipse). In scenario~2 the user virtually grabs the biopsy needle. Two orthogonal projections of the needle axis are displayed on the screen. In scenario~3 the user virtually grabs the biopsy needle. A CT image is projected inside the cadaver.}
    \label{fig:scenarios}
\end{figure}
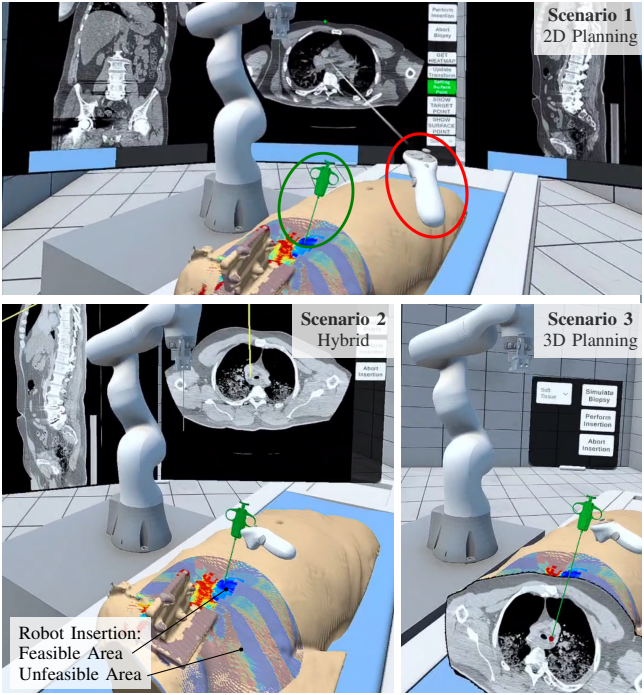
\subsection{Registration and Robot Path Planning}

The transformations of the system are depicted in Fig.~\ref{fig:transformsB}. We perform a hand-eye calibration to estimate the transformation from the robot base coordinate system $B$ to the retro-reflective tracking marker $T\!B$ mounted to the mobile platform. We use the QR24 algorithm \cite{Ernst.2012} to estimate the rigid transform $_{\text{B}}T^{\text{TB}}$ and record 50 robot poses as recommended by the authors. The calibration of the system takes \SI{358(32)}{\second}. We acquire a CT scan of the cadaver and subsequently position the cadaver relative to the robot by adjusting the CT table position. Next, the transform $_{\text{C}}T^{\text{RM}}$ and $_{\text{TB}}T^{\text{C}}$ are recorded with the tracking camera. In the CT image data, the position of the steel balls in the phantom is obtained by thresholding. Additionally, the known CNC machined positions of the grid are matched to the extracted positions using the iterative closest point (ICP) algorithm and the transformation $_{\text{SB}}T^{\text{CT}}$ is estimated. The transformation between the robot base $B$ and the CT imaging system $CT$ is then estimated with

\begin{align}
    _{\text{B}}T^{\text{CT}} = {}_{\text{B}}T^{\text{TB}}\,{}_{\text{TB}}T^{\text{C}}\, _{\text{C}}T^{\text{RM}}\, _{\text{RM}}T^{\text{SB}}\, _{\text{SB}}T^{\text{CT}}.
\end{align}

The communication links and protocols are depicted in Fig.~\ref{fig:systemCommunication}. We use a PACS server to transfer CT image data to the workstation in the intervention room. The robot communication is realized with the Robot Operating System (ROS Noetic Ninjemys, Open Robotics, California, USA). We use MoveIt! for robot control \cite{SucanChitta.2012}. This framework provides implementations of general forward and inverse kinematics solvers, path and trajectory planning, including collision avoidance support, and execution of planned paths. For trajectory planning, we define several collision objects: the segmented skin of the cadaver, the CT gantry, and the custom EEF geometry including the mounted needle. We run MoveIt! embedded in a server structure with a TCP communication protocol. To allow a time-efficient feasibility check of insertion paths, we run 100 server instances in parallel. For CT data processing, we utilize 3D~Slicer\footnote{\href{https://www.slicer.org/}{https://www.slicer.org}} and present an open-source module\footnote{\href{https://collaborating.tuhh.de/e-1/robotic_needle_insertion}{https://collaborating.tuhh.de/e-1/robotic\_needle\_insertion}}. We extract skin models and estimate colormaps that include projections of the maximum Hounsfield unit along possible needle insertion paths, indicating soft and hard tissues, as depicted in Fig.~\ref{fig:scenarios}. At discrete points in the colormap, which are spatially separated by \SI{10}{\milli\meter}, we estimate if robot insertions are feasible. The spacing is defined to enable efficient colormap inference (\SI{18.8(2.87)}{\second}) within a reasonable spatial resolution. The user can visually see the dexterity of the robot based on the defined tissue target, body physiology, and the robot base position.

\subsection{VR Needle Path Planning}
We developed three VR scenarios for planning needle insertions that are depicted in Fig.~\ref{fig:scenarios}. All applications are developed in the game engine Unity (Unity ver 6000.0.25.f1, Unity Technologies, California, USA). Each scenario contains a CT scanner, the robot, screens for depicting CT images, and virtual buttons. The user has a hand-held controller for grabbing objects, moving screens, and pressing virtual buttons. A skin model mesh with an embedded heat map depicts the position of the cadaver relative to the robot. We present three planning scenarios that differ in spatial data presentation: 
\\
\textbf{Scenario~1:} The user is presented with three orthogonal CT slices. The respective slice can be defined with a slider at the bottom of the screen. Using the pointer of the hand-held controller, the user can select an insertion point on the depicted CT slice on the screen. A green biopsy needle indicates the planned trajectory on the skin model. The user can visually verify if the desired trajectory is feasible for the robot, i.e., the biopsy needle intersects a field with saturated colors. \\ 
\textbf{Scenario~2:} The user can virtually grab the biopsy needle with the controller and adapt the insertion angle while the biopsy needle tip is fixed at the target point. On the screen, two orthogonal CT scans along the needle axis are depicted, which are updated in real-time. \\
\textbf{Scenario~3:} Similar to scenario~2, the user can adapt the insertion angle of the needle with the controller. A CT scan is projected inside the skin model. The CT scan and the crop of the skin model are updated in real-time during biopsy needle motion. The viewpoint of the CT scan is flipped if the user crosses the transverse plane with the biopsy needle, i.e., the other half of the skin model is shown.\\
In all scenarios, the grey-level mapping of the CT image can be adapted with the controllers. Please note that we do not alter the volumetric data, e.g., apply spatial filters. We aim to preserve interpretable and known-from-routine image features for medical experts.

\subsection{VR Visualization and Interaction}
The VR communication links and protocols are depicted in Fig.~\ref{fig:systemCommunication}. We establish a WiFi Air link between the VR Headset and the VR workstation using the Meta Quest Link-App (ver. 72.0.0.500.353). This communication protocol allows us to perform high-resolution renderings on the GPU (RTX 4090, Nvidia, California, USA) of the workstation. All communication between the VR headset and unity is handled through the OpenXR Interaction Toolkit (ver. 3.0.6.) that allows interaction between the user and the interactables in the scene and tracks the position of the headset $Q3$ in world coordinates $W$. To communicate with ROS nodes in the VR application, we use ROS\#\footnote{\href{https://github.com/siemens/ros-sharp}{https://github.com/siemens/ros-sharp}} (ver. 2.0.0). This allows real-time visualization in the VR application of the robot state, planned robot trajectories, and a video stream provided by a camera positioned inside the intervention room. The virtual robot is a digital twin of the robot inside the intervention room. The CT images are loaded in DICOM format and converted into a 3D texture. We design shaders that run on a GPU to allow real-time arbitrary slice sampling in CT volumes. We estimate $n$ planes $P_n$ in the CT reference frame. $P_{1-3}$ is defined in scenario~1 by three orthogonal projections along the spatial axis of the CT data. In scenario~2, $P_1$ is defined by the planned needle axis intersecting orthogonally with the sagittal plane and $P_2$ with the axial plane. In scenario~3, $P_1$ is defined by the planned needle axis intersecting orthogonally the sagittal plane. In scenarios 1 and 2 the screens depict the CT pixels located in $P_n$. In scenario~3, image coordinates in $P_1$ are transformed into VR world coordinates $W$ by

\begin{align}
    [x,y,z]^\intercal &= {}_{\text{W}}T^{\text{B}}\,{}_{\text{B}}T^{\text{CT}}\,_{\text{CT}}T^{\text{P1}}[\hat{x},\hat{y},0]^\intercal 
\end{align} 

with the world coordinates $[x,y,z]$, in-plane coordinates $[\hat{x},\hat{y}]$ and a rigid transformation ${}_{\text{W}}T^{\text{B}}$ between the world and robot base. Texture coordinates located outside the skin mesh are set transparent. The skin model $S\!M$ position is defined as

\begin{align}
    _{\text{W}}T^{\text{SM}} = {}_{\text{W}}T^{\text{B}}\, {}_{\text{B}}T^{\text{CT}}\, {}_{\text{CT}}T^{\text{SM}}
\end{align}

with ${}_{\text{CT}}T^{\text{SM}}$ as the identity matrix. The transformations applied in the VR application are also depicted in Fig.~\ref{fig:transformsA}.

\begin{figure}[t]
        \centering
          \begin{tikzpicture}[every node/.style={inner sep=0,outer sep=0}]
            \node [align=left, anchor=south west] at (0mm,0mm)  {\includegraphics[height = 27mm]{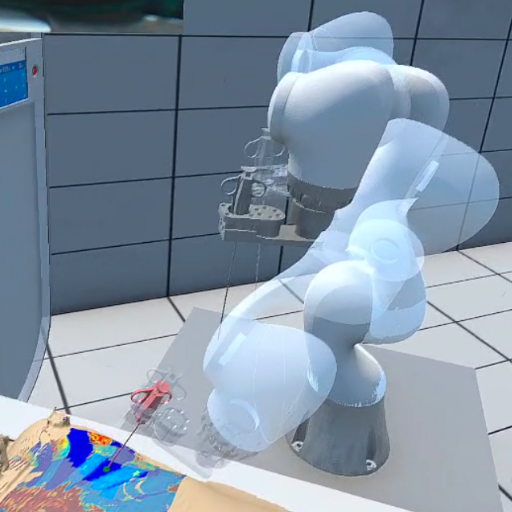}};
            \node [align=left, anchor=south west] at (28mm,0mm)  {\includegraphics[height = 27mm]{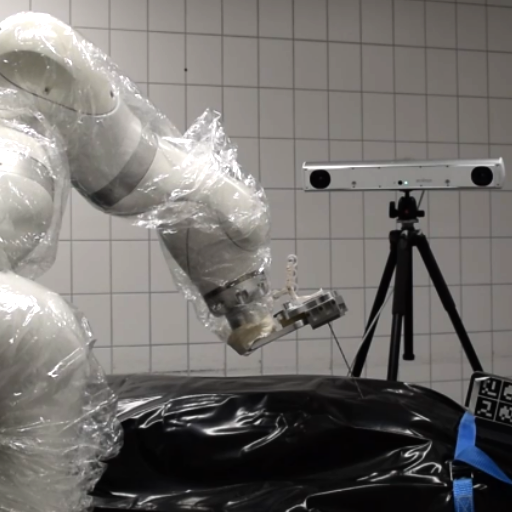}};
            \node [align=left, anchor=south west] at (56mm,0mm)  {\includegraphics[height = 27mm]{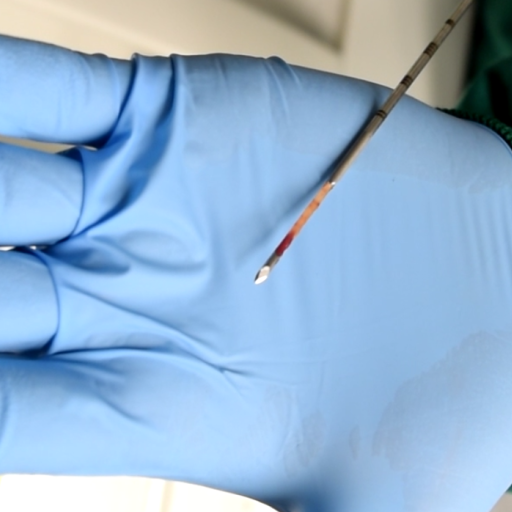}};
            \node [align=left, anchor=south west] at (0mm,28mm)  {\includegraphics[height = 27mm]{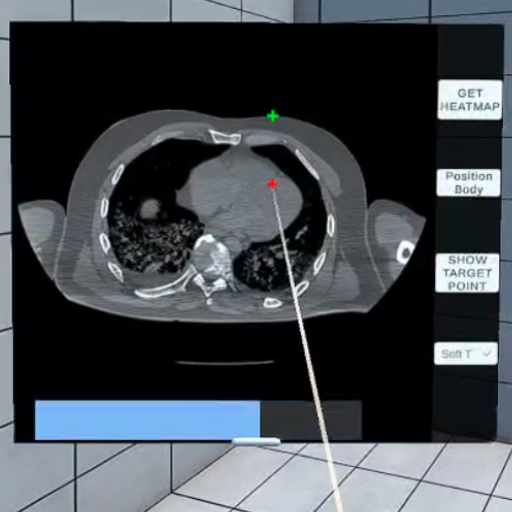}};
            \node [align=left, anchor=south west] at (28mm,28mm)  {\includegraphics[height = 27mm]{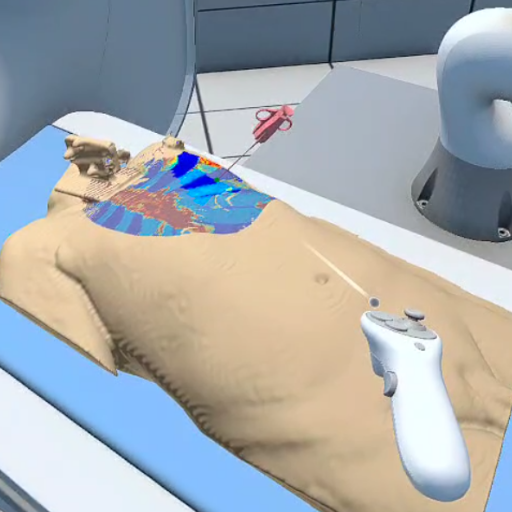}};
            \node [align=left, anchor=south west] at (56mm,28mm)  {\includegraphics[height = 27mm]{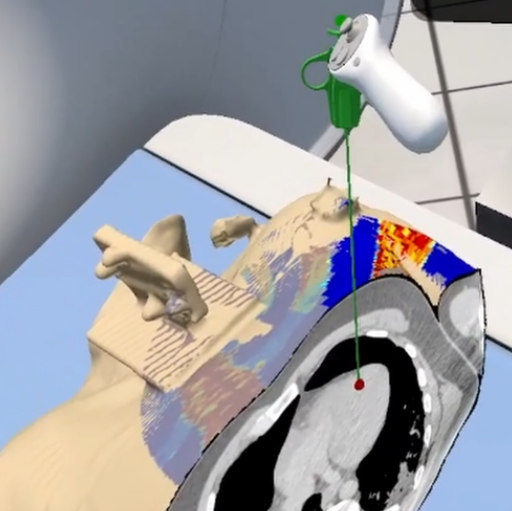}};

            \fill[fill=white, opacity=0.7] (0mm,23mm) rectangle (0mm+27mm,23mm+4mm);
            \fill[fill=white, opacity=0.7] (0mm+28mm,23mm) rectangle (0mm+27mm+28mm,23mm+4mm);
            \fill[fill=white, opacity=0.7] (0mm+28mm+28mm,23mm) rectangle (0mm+27mm+28mm+28mm,23mm+4mm);
            
            \fill[fill=white, opacity=0.7] (0mm,23mm+28mm) rectangle (0mm+27mm,23mm+4mm+28mm);
            \fill[fill=white, opacity=0.7] (0mm+28mm,23mm+28mm) rectangle (0mm+27mm+28mm,23mm+4mm+28mm);
            \fill[fill=white, opacity=0.7] (0mm+28mm+28mm,23mm+28mm) rectangle (0mm+27mm+28mm+28mm,23mm+4mm+28mm);

             \node[align=center] at (14mm,53mm) {\scriptsize 1. Define Target};
             \node[align=center] at (14mm+28mm,53mm) {\scriptsize 2. Estimate Colormap};
             \node[align=center] at (14mm+28mm+28mm,53mm) {\scriptsize 3. Plan Needle Trajectory};
             
             \node[align=center] at (14mm,25mm) {\scriptsize 4. Simulate Robot Motion};
             \node[align=center] at (14mm+28mm,25mm) {\scriptsize 5. Execute Robot Motion};
             \node[align=center] at (14mm+28mm+28mm,25mm) {\scriptsize 6. Extract Tissue Sample};

          \end{tikzpicture}
         \caption{\textbf{Clinical Workflow:} Individual steps during VR assisted needle placement.}
        \label{fig:CW}
\end{figure}
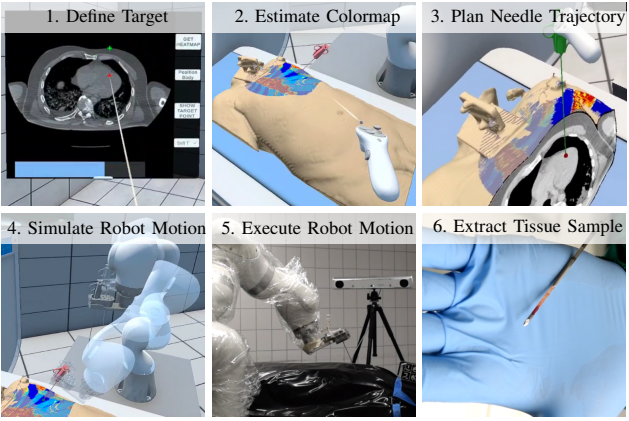

\begin{table}[t]
    \caption{{Clinical Evaluation}}
    \centering
    {\begin{tabular}{cccrccc}
        \toprule
        Cadaver & Gender & Height & Weight & BMI & Age & PM Interval\\
        \midrule
        1 & female & \SI{156}{\centi\meter} & \SI{74.4}{\kilo\gram} & 30.57 & \SI{79}{\year}  &\SI{171}{\hour}\\
        2 & female & \SI{157}{\centi\meter} & \SI{40.5}{\kilo\gram} & 16.43 & \SI{63}{\year}  &\SI{137}{\hour}\\
        3 & male & \SI{191}{\centi\meter} & \SI{101.6}{\kilo\gram} & 27.85 & \SI{55}{\year} &\SI{135}{\hour}\\
          \bottomrule
        \end{tabular}
        }
    \label{tab:Corpse}
\end{table}

\begin{figure*}[t]
    \centering
    \subfloat[]{\includegraphics[height = 41mm,trim=0 0 0 0,clip]{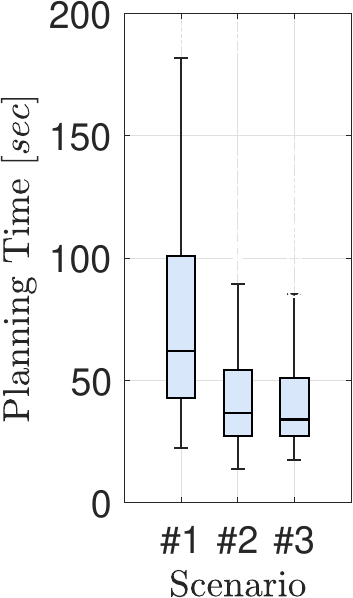}
    \label{fig:userStudyA}}
    \subfloat[]{\includegraphics[height = 41mm,trim=0 0 0mm 0,clip]{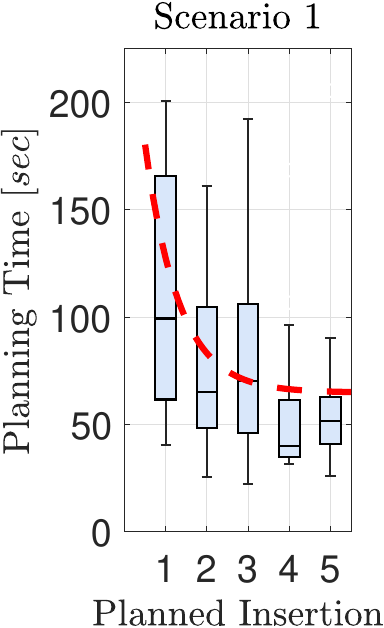}
    \label{fig:userStudyB}}  
    \subfloat[]{\includegraphics[height = 41mm,trim=0 0 0mm 0,clip]{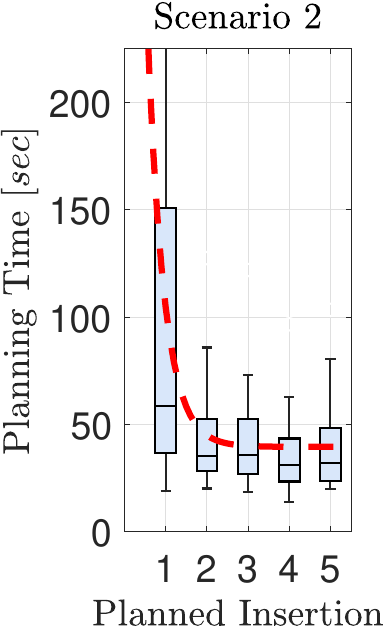}
    \label{fig:userStudyC}}  
    \subfloat[]{\includegraphics[height = 41mm,trim=0 0 0mm 0,clip]{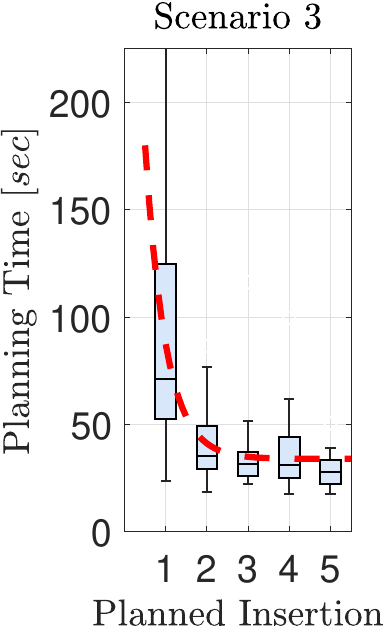}
    \label{fig:userStudyD}} \hfill
    \subfloat[]{\includegraphics[height = 41mm,trim=0 0 0 0,clip]{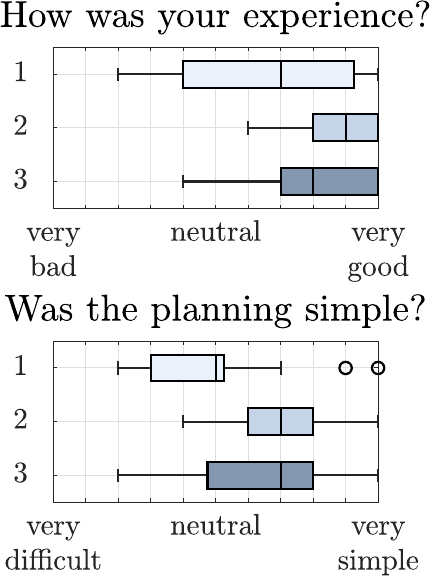}
    \label{fig:userStudyE}}
    \hfill
    \subfloat[]{\includegraphics[height = 41mm,trim=0cm 0 0cm 0,clip]{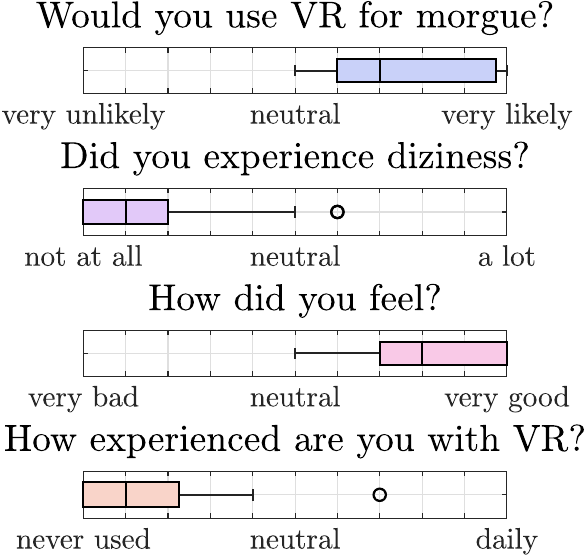}
    \label{fig:userStudyF}} 
    \caption{\textbf{Results usability study:} (a) Planning time for all insertions performed. (b-d) Planning time for subsequent insertions. In red the learning curve is indicated. (e)  The users (n=21) were tasked (e) to rate the individual planning scenarios 1-3 and (f) to rate the overall experience.}
    \label{fig:userStudy}
 \end{figure*}

\subsection{Usability Study}
We evaluate the three VR needle path planning scenarios with a cohort of 11 forensic pathologists and radiologists (\SI{39.55(9.34)}{\year}, 5:5:1, male:female:other) and 10 non-physicians (\SI{27.00(4.22)}{\year}, 9:1:0, male:female:other). Of the forensic pathologists, \SI{45.5}{\percent} had more than \SI{10}{\year} of experience in forensics. Initially, each user had a brief introduction (approx. \SI{5}{\minute}) to the controls and the planning objective. Following, the users were tasked to plan insertions for five predefined tissue targets in the thorax, left kidney, liver, lymph node, pancreas, and spleen. Each participant was given the same CT data and identical predefined targets. The users were tasked to virtually plan needle trajectories that fully avoid bone punctures and are feasible for the robot. The order of performing the scenarios was interchanged between users. Additionally, participants were surveyed to gather insights into their VR experience. The users were tasked to rate their experiences on a numerical Likert scale from 0 to 10. The questionnaire comprised general questions about the interaction with VR and specific questions regarding each planning scenario and clinical acceptance. The scope of the questionnaire is to compare the different planning scenarios.

\subsection{Clinical Evaluation and Workflow}
The clinical workflow during VR assisted needle placement is depicted in Fig.~\ref{fig:CW}. We extract tissue biopsy probes from 3 cadavers with different body types, as indicated in Table~\ref{tab:Corpse}. We defined 16 biopsy targets that are of forensic interest: adrenal gland, coronary artery, heart anterior wall septum, hilum lymph node, kidney, liver, pancreas, prostate, psoas muscle, pulmonary artery, spleen, thoracic aorta, and in total 4 targets in the peripheral as well as central lung regions in the left upper lobe (LUL), left lower lobe (LLL), and right upper lobe (RUL). Tissue probes are extracted with a Gauge 13 biopsy needle system. All needle components are custom manufactured (weLLgo Medical Products GmbH, Wuppertal, Germany). After the robot inserts the hollow guide needle, we first remove the inner co-axial needle. Second, a biopsy needle is inserted through the hollow guide needle. Third, the clamping mechanism of the custom EEF is released. The hollow guide needle remains inside the cadaver, and the robot returns to the defined idle position. Tissue samples are fixed in \SI{4}{\percent} buffered formaldehyde and investigated microscopically for organ and tissue diagnoses in question. For each cadaver and tissue target, we perform needle insertions in all three scenarios leading to a total of 138 performed insertions. Please note, we only extract tissue samples for the first insertion to an annotated target to avoid annotation bias, as previous insertion paths might be visible, and targets can be displaced due to previous needle insertions. In total, 46 tissue probes were extracted. After needle placement, the CT imaging protocol was repeated. For each inserted needle, we manually annotate the tip and the needle base. We define the skin puncture point as the intersection of the annotated needle axis with the skin model. We report the error between the needle tip and the planned target position, the off-axis error defined as the shortest distance between the needle axis and the planned target position, and the distance between the planned and extracted skin puncture position.

\section{Results}

\begin{figure}
   \centering
    \subfloat[]{\includegraphics[height = 50mm,trim=0cm 0cm 0cm 0cm,clip]{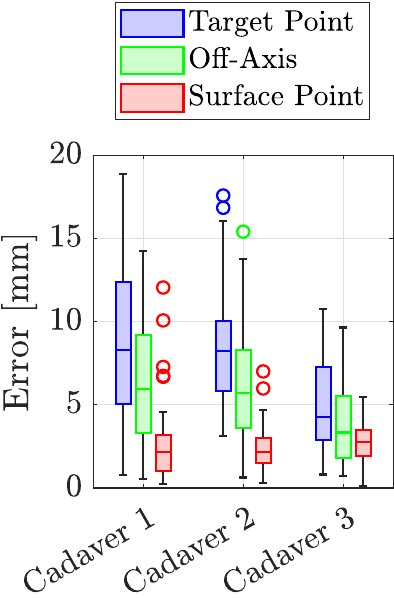}
    \label{fig:histological_resultsA}}
    \subfloat[]{\includegraphics[height = 50mm,trim=0cm 0cm 0cm 0cm,clip]{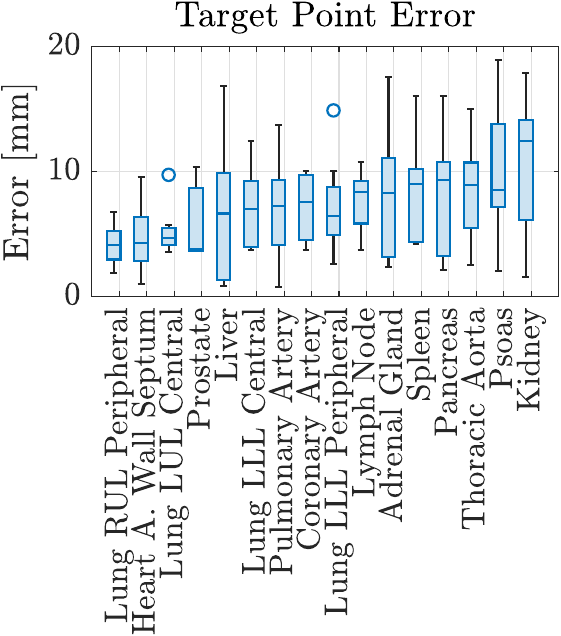}
    \label{fig:histological_resultsB}}

  \caption{\textbf{System Accuracy:} (a) Error of all needles inserted, (b) target error for individual organs.}
\end{figure}

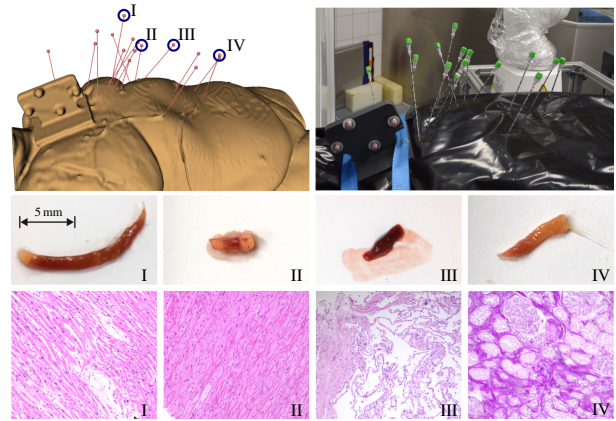
\begin{figure}
\centering
\resizebox{0.46\textwidth}{!}{%
      \begin{tikzpicture}
            \node [align=left, anchor=south west] at (0,0mm)  {\includegraphics[width = 85mm,trim=0mm 0mm 0mm 0mm,clip]{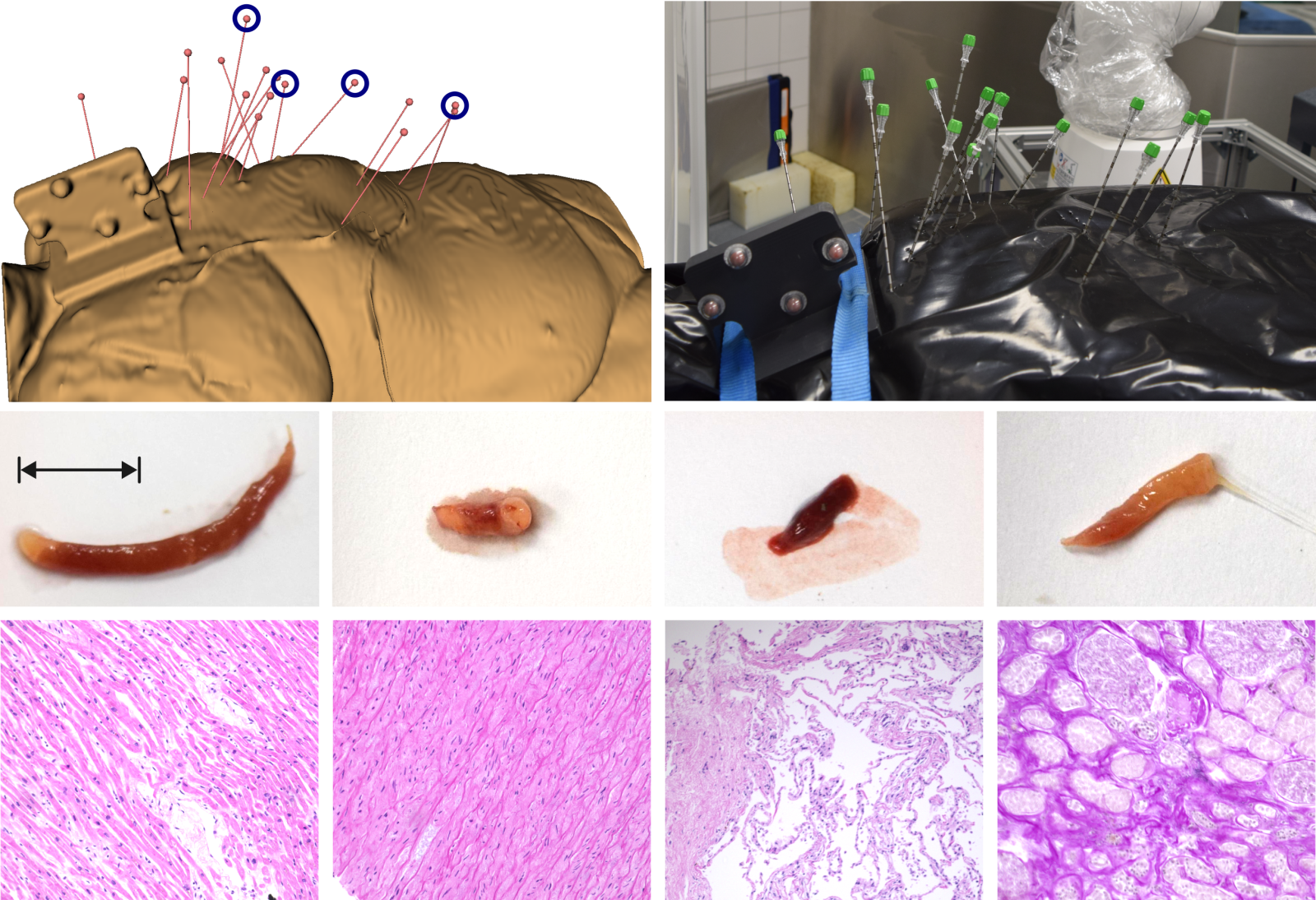}};     

            \node [align=left] at (0.65,3.05){\tiny \SI{5}{\milli\meter}};
           
           \node [align=left] at (1.85,5.9) {\scriptsize{I}};
           \node [align=left] at (2.1,5.55) {\scriptsize{II}};
           \node [align=left] at (2.6,5.55) {\scriptsize{III}};
           \node [align=left] at (3.3,5.4) {\scriptsize{IV}};

           \node [align=left] at (2.0+2.15*0,2.15) {\scriptsize{I}};
           \node [align=left] at (2.0+2.15*1,2.15) {\scriptsize{II}};
           \node [align=left] at (2.0+2.15*2,2.15) {\scriptsize{III}};
           \node [align=left] at (2.0+2.15*3,2.15) {\scriptsize{IV}};
           
           \node [align=left] at (2.0+2.15*0,0.25) {\scriptsize{I}};
           \node [align=left] at (2.0+2.15*1,0.25) {\scriptsize{II}};
           \node [align=left] at (2.0+2.15*2,0.25) {\scriptsize{III}};
           \node [align=left] at (2.0+2.15*3,0.25) {\scriptsize{IV}};

        \end{tikzpicture}
        }
  \caption{\textbf{Histopathological Evaluation:} example tissue biopsy samples and histopathological images; (I) heart wall anterior septum, (II) thoracic aorta, (III) lung, and (IV) kidney.}
    \label{fig:histological_results}
\end{figure}

\subsection{Usability study}
The results of the usability study are depicted in Fig.~\ref{fig:userStudy}. As shown in Fig.~\ref{fig:userStudyA}, the least amount of planning time was required in scenarios 2 and 3 with \SI{53.76(53.51)}{\second} and \SI{46.47(34.00)}{\second}, respectively. It stands out that scenario~1 was the least time-efficient with a mean planning time of \SI{82.15(61.41)}{\second}. The learning curves, depicted in Fig.~\ref{fig:userStudyB}-\ref{fig:userStudyD}, show a reduction in planning time of \SI{53}{\percent}, \SI{38}{\percent}, and \SI{33}{\percent} for scenarios 1-3, respectively. After performing four insertions in each scenario, the users were able to plan the subsequent needle insertion in \SI{67.57(57.20)}{\second}, \SI{40.16(22.03)}{\second}, and \SI{28.74(8.28)}{\second} for scenarios 1-3, respectively. The questionnaire evaluation is depicted in Fig.~\ref{fig:userStudyE}–\ref{fig:userStudyF}. The majority of users did not have any experience with VR prior. Also, the majority of users did not experience any dizziness or motion sickness in the VR application. The users felt positively comfortable during planning. Forensic experts report a positive tendency to use the planning application during clinical routine with \SI{7.55(1.97)}{}. Considering the respective scenarios, depicted in Fig.~\ref{fig:userStudyF}, users experienced the simplest planning approach in scenario~2 with \SI{7.00(1.64)}{}. Scenario~1 was considered the most difficult for planning with \SI{4.86(2.03)}{}, while scenario~3 was in between with \SI{6.43(2.09)}{}. A similar tendency can be observed considering the user's experience with the most positive being scenario~2 with \SI{8.71(1.52)}{}, followed by scenario~3 with \SI{8.05(1.60)}{}, and scenario~1 with \SI{6.67(2.78)}{}.

\subsection{Post Mortem Evaluation}
We inserted 138 coaxial needles in 3 cadavers. Five needles were removed from the study, as they did not stay fixed after insertion, based on the shallow needle insertion depth due to the low Body Mass Index (BMI) of cadaver 2. Also, one needle was removed from the study evaluation, as it was touched by the robot EEF during the insertion of a subsequent needle. We estimate a colormap for each target and check on average \SI{1649(365)}{} needle insertion paths. The target, off-axis, and surface point errors are depicted in the boxplot in Fig.~\ref{fig:histological_resultsA}. We report a mean target error of \SI{7.35(4.10)}{\milli\meter}, a mean off-axis error of \SI{5.30(3.25)}{\milli\meter}, and a mean surface point error of \SI{2.62(1.76)}{\milli\meter}. It stands out that only minor deviations are observed for the surface point error between all cadavers, while a slightly larger deviation is observed for the target point error. Further, we report the target point error across all organ targets in Fig.~\ref{fig:histological_resultsB}. From each cadaver we extract a set of tissues. We report a system latency between the real robot and the digital twin of \SI{50.25(0.75)}{\milli\second}. Tissue probes and histopathological scans are depicted in Fig.~\ref{fig:histological_results}. Histopathological analysis confirmed \SI{65}{{\percent}} of samples with the majority of misses reported for small structures, e.g., the coronary artery. 

\section{Discussion and Conclusion}
We present a robotic system for extracting tissue samples safely from potentially infectious cadavers. Needle trajectory planning and execution are performed remotely in a custom VR application, physically separated from the cadaver. The VR application allows efficient 3D needle insertion planning and enables users without robotics expertise to intuitively take into account the robot's dexterity. The system can be operated over a large distance between physician and cadaver, considering latencies, bandwidth, and packet loss. The lightweight robotic arm is mounted on a cart which allows fast deployment at a medical center and a high degree of flexibility in contrast to fixed ceiling-mounted systems \cite{Ebert.2010}.

We study 3 scenarios for needle path planning. The majority of users reported a positive VR experience. All users were able to plan needle insertions after a short introduction to the controls. The learning curves of all scenarios show an initial steep decrease and an overall planning time reduction of \SIrange{33}{53}{\percent}. No strong deviations are observed between physicians and non-physicians for all scenarios. This indicates that our needle path planning is simple and intuitive. Future studies on immersiveness and long-term user experience are planned. Also, needle insertions are time efficient. Scenario~3 stands out with an average planning time of \SI{28.74(8.28)}{\second}, including inference time of the colormap. The most positive experience and simplest planning approach was given in scenario~2, based on the responses in the questionnaire. In contrast, users report scenario~1 as considerably more complex and rate the experience more negatively. Accordingly, planning time in scenario~1 is approximately twofold greater compared to scenario~3. This is particularly interesting as the design of scenario~1 is based on a conventional planning approach.

With the proposed system, we performed in total 132 needle insertions in three cadavers. We report a similar target point error as Neidhardt et al. \cite{neidhardt2022robotic} with  \SI{7.19(4.22)}{\milli\meter} compared to our mean target point error of \SI{7.35(4.10)}{\milli\meter}. It stands out that the surface point error is consistently lower with \SI{2.62(1.76)}{\milli\meter}. This might be related to the movement of the needles after detachment from the robot’s EEF. This is especially apparent during very shallow insertions, e.g. in cadaver 2 due to the low BMI. Additionally, during robotic needle placement, \SI{15}{\milli\meter} is subtracted from the insertion depth to position the biopsy punch at the target center. During accuracy evaluation, the positioned needle is virtually extended by \SI{15}{\milli\meter}, making it sensitive to small orientation deviations, such as those occurring after detachment from the robot. Also, note that the target error is less in cadaver 3 with \SI{5.09(2.84)}{\milli\meter}, which can be related to the overall deeper needle insertions and thereby less movement of the needle after detachment. We histopathologically confirm \SI{65}{{\percent}} of samples. The success rate is dependent on (1) the medical biopsy needle, which might not be ideal for all tissue types and consistencies, (2) small target volumes, e.g., the coronary arteries have a maximum diameter of \SI{4.16}{\milli\meter} \cite{muneeb2023assessment}, and (3) the degree of autolysis degrading biopsy quality. Unlike manual ultrasound-guided methods that rely on physician expertise, the robot ensures consistent needle placement regardless of user experience. This makes it well-suited for systematic sampling, e.g., in the context of a multicentre national autopsy registry \cite{von2022first}. Our system could support more effective public health responses in future pandemics by reducing exposure risk during cause-of-death determination and tissue sampling.

\ifCLASSOPTIONcaptionsoff
  \newpage
\fi

\bibliographystyle{IEEEtran}
\bibliography{refs}

\begin{thebibliography}{10}
\providecommand{\url}[1]{#1}
\csname url@samestyle\endcsname
\providecommand{\newblock}{\relax}
\providecommand{\bibinfo}[2]{#2}
\providecommand{\BIBentrySTDinterwordspacing}{\spaceskip=0pt\relax}
\providecommand{\BIBentryALTinterwordstretchfactor}{4}
\providecommand{\BIBentryALTinterwordspacing}{\spaceskip=\fontdimen2\font plus
\BIBentryALTinterwordstretchfactor\fontdimen3\font minus \fontdimen4\font\relax}
\providecommand{\BIBforeignlanguage}[2]{{%
\expandafter\ifx\csname l@#1\endcsname\relax
\typeout{** WARNING: IEEEtran.bst: No hyphenation pattern has been}%
\typeout{** loaded for the language `#1'. Using the pattern for}%
\typeout{** the default language instead.}%
\else
\language=\csname l@#1\endcsname
\fi
#2}}
\providecommand{\BIBdecl}{\relax}
\BIBdecl

\bibitem{brandner2022contamination}
J.~M. Brandner \emph{et~al.}, ``Contamination of personal protective equipment during covid-19 autopsies,'' \emph{Virchows Archiv}, vol. 480, no.~3, pp. 519--528, 2022.

\bibitem{lahmer2024postmortem}
T.~Lahmer \emph{et~al.}, ``Postmortem minimally invasive autopsy in critically ill covid-19 patients at the bedside: A proof-of-concept study at the icu,'' \emph{Diagnostics}, vol.~14, no.~3, p. 294, 2024.

\bibitem{terence2023minimally}
A.~Terence~Azeke \emph{et~al.}, ``Minimally invasive tissue sampling via post mortem ultrasound: A feasible tool (not only) in infectious diseases—a case report,'' \emph{Diagnostics}, vol.~13, no.~16, p. 2643, 2023.

\bibitem{levy2021clinical}
S.~Levy \emph{et~al.}, ``Clinical evaluation of a robotic system for precise ct-guided percutaneous procedures,'' \emph{Abdominal radiology}, vol.~46, no.~10, pp. 5007--5016, 2021.

\bibitem{bodard2023percutaneous}
S.~Bodard \emph{et~al.}, ``Percutaneous liver interventions with robotic systems: a systematic review of available clinical solutions,'' \emph{Brit. J. Radiol.}, vol.~96, no. 1152, p. 20230620, 2023.

\bibitem{moser2013novel}
C.~Moser \emph{et~al.}, ``A novel laser navigation system reduces radiation exposure and improves accuracy and workflow of ct-guided spinal interventions: a prospective, randomized, controlled, clinical trial in comparison to conventional freehand puncture,'' \emph{European Journal of Radiology}, vol.~82, no.~4, pp. 627--632, 2013.

\bibitem{mieling2023collaborative}
R.~Mieling \emph{et~al.}, ``Collaborative robotic biopsy with trajectory guidance and needle tip force feedback,'' in \emph{2023 IEEE International Conference on Robotics and Automation (ICRA)}.\hskip 1em plus 0.5em minus 0.4em\relax IEEE, 2023, pp. 6893--6900.

\bibitem{Ebert.2010}
L.~C. Ebert \emph{et~al.}, ``Virtobot--a multi-functional robotic system for 3d surface scanning and automatic post mortem biopsy,'' \emph{Int J Med Robot Comput Assist Surg}, vol.~6, no.~1, pp. 18--27, 2010.

\bibitem{neidhardt2022robotic}
M.~Neidhardt \emph{et~al.}, ``Robotic tissue sampling for safe post-mortem biopsy in infectious corpses,'' \emph{IEEE T-MRB}, vol.~4, pp. 94--105, 2022.

\bibitem{evans2025augmented}
M.~Evans \emph{et~al.}, ``Augmented reality for surgical navigation: A review of advanced needle guidance systems for percutaneous tumor ablation,'' \emph{Radiology: Imaging Cancer}, vol.~7, no.~1, p. e230154, 2025.

\bibitem{murphy2024diagnostic}
P.~M. Murphy \emph{et~al.}, ``Diagnostic performance of a next-generation virtual/augmented reality headset: A pilot study of diverticulitis on ct,'' \emph{JIIM}, pp. 1--8, 2024.

\bibitem{pinter2020slicervr}
C.~Pinter \emph{et~al.}, ``Slicervr for medical intervention training and planning in immersive virtual reality,'' \emph{IEEE T-MRB}, vol.~2, no.~2, pp. 108--117, 2020.

\bibitem{Ernst.2012}
F.~Ernst \emph{et~al.}, ``Non-orthogonal tool/flange and robot/world calibration,'' \emph{Int J Med Robot Comput Assist Surg}, vol.~8, no.~4, pp. 407--420, 2012.

\bibitem{SucanChitta.2012}
\BIBentryALTinterwordspacing
I.~A. Sucan and S.~Chitta, ``Moveit,'' 2012. [Online]. Available: \url{moveit.ros.org.}
\BIBentrySTDinterwordspacing

\bibitem{muneeb2023assessment}
M.~Muneeb \emph{et~al.}, ``Assessment of the dimensions of coronary arteries for the manifestation of coronary artery disease,'' \emph{Cureus}, vol.~15, 2023.

\bibitem{von2022first}
S.~von Stillfried \emph{et~al.}, ``First report from the german covid-19 autopsy registry,'' \emph{The Lancet Regional Health--Europe}, vol.~15, 2022.

\end{thebibliography}

\newpage

\end{document}